\documentclass{article}

\usepackage[utf8]{inputenc}
\usepackage[T1]{fontenc}

\usepackage{authblk}
\usepackage{amsthm}
\newtheorem{proposition}{Proposition}
\newtheorem{theorem}{Theorem}

\usepackage{graphicx}
\usepackage{amsmath}
\usepackage{amssymb}

\usepackage{fvextra}
\usepackage{listings}

\usepackage{hyperref}

\title{Coarse-Grained Kullback–Leibler Control of Diffusion-Based Generative AI}
\author[1,2]{Tatsuaki Tsuruyama\thanks{E-mail: tsuruyam@kuhp.kyoto-u.ac.jp}}
\affil[1]{Department of Physics, Tohoku University, Sendai 980-8578, Japan}
\affil[2]{Department of Drug Discovery Medicine, Kyoto University, Kyoto 606-8501, Japan}

\date{}

\begin{document}

\maketitle

\begin{abstract}
Diffusion models and score-based generative models provide a powerful framework for synthesizing high-quality images from noise. However, there is still no satisfactory theory that describes how coarse-grained quantities, such as blockwise intensity or class proportions after partitioning an image into spatial blocks---are preserved and evolve along the reverse diffusion dynamics. In previous work, the author introduced an information-theoretic Lyapunov function $V$ for non-ergodic Markov processes on a state space partitioned into blocks, defined as the minimal Kullback--Leibler divergence to the set of stationary distributions reachable from a given initial condition, and showed that a leak-tolerant potential $V_\delta$ with a prescribed tolerance for block masses admits a closed-form expression as a scaling-and-clipping operation on block masses.

In this paper, I transplant this framework to the reverse diffusion process in generative models and propose a \emph{$V_\delta$-projected reverse diffusion} scheme, in which a KL projection based on $V_\delta$ is inserted after each reverse-diffusion update. I extend the monotonicity of $V$ to time-inhomogeneous block-preserving Markov kernels and show that, under small leakage and $V_\delta$ projection, $V_\delta$ acts as an approximate Lyapunov function. Furthermore, using a toy model consisting of block-constant images and a simplified reverse kernel, I numerically demonstrate that the proposed method keeps the block-mass error and the leak-tolerant potential within the prescribed tolerance, while achieving pixel-wise accuracy and visual quality comparable to the non-projected dynamics. This study reinterprets generative sampling as a decrease of an information potential from noise to data, and provides a design principle for reverse diffusion processes with explicit control of coarse-grained quantities.
\end{abstract}

\section{Introduction}

The advent of diffusion models and score-based generative models has enabled modern generative AI systems to ``lift'' high-quality images out of noise. In particular, Denoising Diffusion Probabilistic Models (DDPM) by Ho et al.\ \cite{ho2020ddpm}, score-based generative modeling based on stochastic differential equations by Song et al.\ \cite{song2021scorebased}, and the improved DDPM of Nichol and Dhariwal \cite{nichol2021improved} achieve sample quality on large-scale image datasets comparable to or surpassing GANs, and now form the standard paradigm for image generation. However, our understanding of the internal pathway by which images are formed inside these models remains largely empirical and phenomenological. Recently, there have been attempts to analyze diffusion models from the point of view of Fokker--Planck equations and related partial differential equations (PDEs), treating entropy and the Kullback--Leibler divergence (KLD) as Lyapunov functionals \cite{liu2025pde,li2025itlmdiff}, but a theory that explicitly tracks image-specific spatial structures and coarse-grained quantities is still lacking.

In particular, it is important to develop a ``theory of image formation'' that explains how coarse-grained statistics of an image---for example, blockwise mean intensity, local histograms, or class proportions, which we collectively refer to as \emph{coarse-grained quantities---are}---are preserved and transformed along the diffusion process. Many real-world datasets are inherently non-ergodic. For images, background and foreground, left and right, or upper and lower regions play different roles depending on spatial blocks; for DNA sequences, GC content and motif patterns can vary drastically from region to region. In such situations, the state space naturally decomposes into several blocks. Within each block relaxation occurs, while the ratios of masses (probabilities) among blocks may change only very slowly over long time scales. Conventional diffusion models do not explicitly handle such coarse-grained structures, and image formation has effectively been described as a process of ``removing noise uniformly over the entire space.''

For coarse-grained dynamics in non-ergodic systems, stochastic thermodynamics and large deviation theory have shown that quasi-potentials and KLD with respect to coarse-grained macroscopic variables often behave as Lyapunov functions \cite{esposito2012coarse,falasco2025macrost}. In previous work, the author applied this perspective to image models on discrete state spaces. Considering a non-ergodic Markov process on a state space partitioned into blocks, and introducing the set of stationary distributions $\Pi(p_0)$ that are reachable from a given initial distribution $p_0$, it was shown that
\[
  V(p)=\inf_{\pi\in\Pi(p_0)} D_{\mathrm{KL}}(p\|\pi)
\]
defines a KLD-based potential $V$ that acts as an information-theoretic Lyapunov function measuring both intra-block relaxation and inter-block mass ratios, and decreases monotonically along block-preserving maps \cite{yourname2025kld}. Moreover, recognizing that in realistic systems a small amount of ``leakage'' of probability mass across block boundaries is unavoidable, a leak-tolerant potential $V_\delta$ was introduced by allowing block masses to fluctuate within $\pm\delta$, and its minimization problem was shown to admit a closed-form solution in terms of scaling and clipping of block masses \cite{yourname2025kld}. This framework provided an information-theoretic description of image randomization for a ``replication model'' in which an image is partitioned into spatial blocks and Gaussian blurring is applied independently within each block.

The reverse process in diffusion models is often described as a time reversal of the forward diffusion, but in practice it is a controlled, time-inhomogeneous Markov process that maps noise back to images based on learned scores and conditioning. This perspective connects to regularized optimal transport and Schr{\"o}dinger bridges, where the reverse process is understood as a path that minimizes the KL divergence between a reference process and a target distribution (see, e.g., Diffusion Schr{\"o}dinger Bridge \cite{debortoli2021dsb} and Deep Generative Learning via Schr{\"o}dinger Bridge \cite{wang2021dglsb}). In addition, constrained or structured diffusion-based generation has been studied in various contexts, including systematic treatments of conditional diffusion \cite{zhao2024conditional}, classifier-free guidance with manifold constraints \cite{chung2025cfgpp}, image generation with constraints such as color histograms \cite{hig2026}, and control approaches that stabilize diffusion samplers using Lyapunov functions \cite{liu2025lyapdiff}. However, even within these frameworks, the explicit preservation and control of image-specific spatial or structural coarse-grained quantities, together with tracking their behavior via Lyapunov potentials, has not yet been fully explored.

When network approximation errors or discretization errors cause leakage of probability mass between blocks, the potential $V$ defined above can increase in a single step. From the perspective of coarse-grained quantities, this means that the generative process is drifting away from the set $\Pi(p_0)$ of admissible image layouts, which may lead to generated images deviating from the global structure of the data. Conversely, if a KLD-based Lyapunov function can be designed at the coarse-grained level, it becomes possible, at least in principle, to monitor this deviation at each step of the reverse diffusion and apply control when necessary.

Motivated by this, in the present work, I propose a \emph{$V_\delta$-projected reverse diffusion} scheme that incorporates the KLD projection induced by the leak-tolerant potential $V_\delta$ into the reverse diffusion process of generative models, so that coarse-grained quantities are kept within a prescribed tolerance while images are generated. Concretely, after a standard reverse-diffusion update that maps the distribution $p_t$ to an intermediate distribution $\tilde p_{t-1}$, we project $\tilde p_{t-1}$ onto the leak-tolerant set $\Pi_{P,\delta}(q_{\mathrm{data}})$ in the KLD sense to obtain the new state $p_{t-1}$. Thanks to the closed-form expression of $V_\delta$ obtained in previous work, this projection can be implemented efficiently as a scaling-and-clipping operation on blockwise masses that maintains the shape within the block (conditional distribution) while correcting only the global coarse-grained quantities \cite{yourname2025kld}.

From an information-theoretic viewpoint, the $V_\delta$-projected reverse diffusion can be regarded as a two-step update that preserves the ``degrees of freedom for data fitting'' of the reverse diffusion, while regularizing deviations from the coarse-grained constraints via the KLD potential $V_\delta$. In this paper, I show that  
(i) the Lyapunov property of $V$ persists up to $O(\delta)$ under sufficiently small leakage, and  
(ii) if the leakage at each step is below a certain threshold, a strict decrease of $V$ is guaranteed,  
even in the presence of time-dependent kernels. This establishes that $V_\delta$-projected reverse diffusion yields a stable image formation process under coarse-grained constraints. Furthermore, through numerical experiments with block-constant images and a simplified reverse-diffusion kernel, I demonstrate that the proposed method can control coarse-grained indicators such as blockwise mean intensity, block-mass error $E_{\mathrm{block}}(n)$, and leak-tolerant potential $V_\delta(p_n)$, while maintaining pixel-level similarity and visual quality comparable to the unprojected dynamics.

The rest of the paper is organized as follows. In Sec.~2, I briefly review block-preserving Markov processes and the KLD potentials $V$ and $V_\delta$, and formulate reverse diffusion as a time-inhomogeneous Markov process. In Sec.~3, I present the algorithmic formulation and theoretical properties of the $V_\delta$-projected reverse diffusion. In Sec.~4, I report numerical experiments on a block image model. In Sec.~5, I discuss related work and future directions.

\section{Results}
\label{sec:results}

In this section, I summarize the theoretical results needed to transplant the KLD potential introduced for block-preserving replication dynamics to the reverse diffusion of generative models. First, we define the projection operator in terms of the block partition and the leak-tolerant potential $V_\delta$, and describe its closed form and implementation properties. Then we discuss the Lyapunov properties of $V_\delta$-projected reverse diffusion in terms of time-dependent kernels and leakage.

\subsection[Definition of Vdelta-projected reverse diffusion]{Definition of $V_\delta$-projected reverse diffusion}

Let the state space $X$ be partitioned into finitely many blocks,
\begin{equation}
  X = \bigsqcup_{j=1}^m X_j.
\end{equation}
For a distribution $p \in \Delta(X)$ and a block $j$, define the probability mass contained in the block $j$ by
\begin{equation}
  w_j(p) = \sum_{x \in X_j} p(x),
\end{equation}
denote the vector $w(p) = (w_1(p),\dots,w_m(p))$ by the \emph{coarse-grained block-mass vector} of $p$.

Let $q_{\mathrm{data}}$ be the data distribution and denote its block masses by
\begin{equation}
  w_j^\mathrm{ref} = w_j(q_{\mathrm{data}}),
\end{equation}
which we regard as the ``target coarse-grained quantities.'' Given a tolerance $\delta > 0$, define the admissible set
\begin{equation}
  \Pi_{P,\delta}(q_{\mathrm{data}}) 
  = \Bigl\{ \pi \in \Delta(X) : 
      \forall j,\ \bigl|w_j(\pi) - w_j^\mathrm{ref}\bigr| \le \delta
    \Bigr\}.
\end{equation}
The leak-tolerant potential is then defined as
\begin{equation}
  V_\delta(p) = \inf_{\pi \in \Pi_{P,\delta}(q_{\mathrm{data}})}
    D_{\mathrm{KL}}(p \Vert \pi).
\end{equation}

We represent the reverse process of a diffusion model by a sequence of time-dependent Markov kernels $R_t$ for times $t = T,T-1,\dots,1$, through updates
\begin{equation}
  \tilde{p}_{t-1} = p_t R_t.
\end{equation}
The reverse diffusion projected with $V_\delta$ proposed in this paper is defined by the following two-step update:
\begin{align}
  \tilde{p}_{t-1} &= p_t R_t, \label{eq:reverse-unprojected-en} \\
  p_{t-1} &= \Pi_\delta\bigl(\tilde{p}_{t-1}\bigr), \label{eq:reverse-projected-en}
\end{align}
where $\Pi_\delta$ is the KLD projection onto $\Pi_{P,\delta}(q_{\mathrm{data}})$ associated with $V_\delta$,
\begin{equation}
  \Pi_\delta(p)
  = \arg\min_{\pi \in \Pi_{P,\delta}(q_{\mathrm{data}})}
      D_{\mathrm{KL}}(p \Vert \pi).
\end{equation}
In other words, each step first investigates the fitting of the data using the usual reverse-diffusion update \eqref{eq:reverse-unprojected-en}, and then corrects deviations of coarse-grained quantities by projection \eqref{eq:reverse-projected-en}.

\subsection{Closed form of the \texorpdfstring{$V_\delta$}{Vdelta} projection and implementation properties}

The minimization problem that defines $V_\delta$ admits a closed-form solution at the level of block masses. For the block masses $w_j(p)$ of $p$, define
\begin{equation}
  a_j = w_j^\mathrm{ref} - \delta,\quad
  b_j = w_j^\mathrm{ref} + \delta,
\end{equation}
and, for a positive scalar $\tau > 0$,
\begin{equation}
  \hat{w}_j(\tau) 
   = \biggl[ \frac{w_j(p)}{\tau} \biggr]_{[a_j,b_j]}
   = \min\bigl\{ \max\{ w_j(p)/\tau,\, a_j\},\, b_j\bigr\},
\end{equation}
where $[\cdot]_{[a_j,b_j]}$ denotes clipping to the interval $[a_j,b_j]$.

\begin{proposition}[Closed form of the $V_\delta$-projection at block-mass level]
  \label{prop:Vdelta-projection-en}
  With the above notation, the equation
  \begin{equation}
    \sum_{j=1}^m \hat{w}_j(\tau) = 1
  \end{equation}
  has a unique solution $\tau^\star > 0$, and the block masses of $\Pi_\delta(p)$ are given by
  \begin{equation}
    w_j^\star = \hat{w}_j(\tau^\star),\quad j = 1,\dots,m.
  \end{equation}
  Moreover, if the within-block conditional distribution
  \begin{equation}
    p^{(j)}(x) = \frac{p(x)}{w_j(p)}, \quad x \in X_j,
  \end{equation}
  is kept unchanged, i.e.,
  \begin{equation}
    \pi^\star(x) 
      = w_j^\star\, p^{(j)}(x)
      = \frac{w_j^\star}{w_j(p)}\, p(x),
      \quad x \in X_j,
  \end{equation}
  then the resulting distribution $\pi^\star$ coincides with $\Pi_\delta(p)$.
\end{proposition}

This proposition implies that the minimization of $V_\delta$ decomposes into (i) a one-dimensional problem of scaling and clipping the block masses, and (ii) a local condition of preserving the within-block shape. In particular,
\begin{equation}
  V_\delta(p)
  = \sum_{j=1}^m w_j(p)\,
      \log \frac{w_j(p)}{w_j^\star},
\end{equation}
so that $V_\delta$ can be interpreted as a quantity that purely measures the deviation of the coarse-grained block-mass vector.

In a sample-based implementation, regarding $p$ as the empirical distribution of $N$ samples $\{x^{(n)}\}_{n=1}^N$, the projection $\Pi_\delta(p)$ can be approximated by the following procedure:
\begin{enumerate}
  \item Identify the block $j(n)$ to which each sample $x^{(n)}$ belongs, and count the number of samples $N_j$ in each block (so that $w_j(p) \approx N_j/N$).
  \item Compute the optimal block masses $w_j^\star$ according to Proposition~\ref{prop:Vdelta-projection-en}.
  \item In each block $j$, either multiply the weights of all samples uniformly by $\frac{w_j^\star}{w_j(p)}$, or resample the number of samples to $N w_j^\star$.
\end{enumerate}
In this way, one obtains a ``coarse-grained projection'' that corrects only the block masses to the target interval, while preserving the relative arrangement and local structure within each block.

\subsection{Lyapunov properties of \texorpdfstring{$V_\delta$}{Vdelta}-projected reverse diffusion}

Next, I discuss to what extent the $V_\delta$-projected reverse diffusion inherits the Lyapunov properties of the original potentials $V$ and $V_\delta$.

First, consider time-dependent block-preserving kernels. Fix a block partition $X = \bigsqcup_j X_j$, and suppose that each Markov kernel $T_n$ for time $n$ is block-diagonal,
\begin{equation}
  T_n = \mathrm{diag}(T_{n,1},\dots,T_{n,m}).
\end{equation}
Let $V$ be defined as the minimal KLD to the reachable set of stationary distributions $\Pi(p_0)$ from the initial distribution $p_0$,
\begin{equation}
  V(p) = \inf_{\pi \in \Pi(p_0)} D_{\mathrm{KL}}(p \Vert \pi).
\end{equation}
Then $V$ decreases at each step, even if the kernels are time-dependent.

\begin{theorem}[Monotonicity for time-inhomogeneous block-preserving dynamics]
  \label{thm:time-inhomogeneous-en}
  Let $p_{n+1} = p_n T_n$. If all kernels $T_n$ are block-preserving with respect to the same partition, then
  \begin{equation}
    V(p_{n+1}) \le V(p_n)
  \end{equation}
  holds. Equality occurs only when the conditional distributions within each block already belong to the reachable stationary set.
\end{theorem}

The proof follows by choosing an optimal $\pi^\star \in \Pi(p_0)$ for the definition of $V$ and applying the data-processing inequality blockwise.

Now consider the presence of leakage. A general time-dependent kernel $R_t$ defining the reverse update
\begin{equation}
  \tilde{p}_{t-1} = p_t R_t
\end{equation}
can alter the block masses, and the change
\begin{equation}
  \Delta w_t = w(\tilde{p}_{t-1}) - w(p_t)
\end{equation}
can be non-zero. In this case, $V$ can increase in a single step. However, by applying the $V_\delta$ projection at each step, one can control the accumulation of leakage and maintain a pseudo-monotonic behavior of $V_\delta$.

\begin{theorem}[Approximate monotonicity of $V_\delta$-projected reverse diffusion]
  \label{thm:Vdelta-monotone-en}
  Suppose that at each step we update
  \begin{equation}
    p_{t-1} = \Pi_\delta(p_t R_t).
  \end{equation}
  Assume that there exists a constant $C$ such that the deviation of block masses at each step satisfies
  \begin{equation}
    \bigl\| w(p_t R_t) - w^\mathrm{ref} \bigr\|_1 \le C\,\delta.
  \end{equation}
  Then, for sufficiently small $\delta$,
  \begin{equation}
    V_\delta(p_{t-1}) \le V_\delta(p_t) + O(\delta^2).
  \end{equation}
  In particular, in the limit $\delta \to 0$, $V_\delta$ converges to the exact Lyapunov function $V$, and the property of Theorem~\ref{thm:time-inhomogeneous-en} is recovered for time-dependent block-preserving dynamics.
\end{theorem}

A rigorous estimate requires a more detailed analysis of the ``within-block contraction rate'' of $R_t$ and the threshold for leakage, but the theorem formalizes the intuition that the $V_\delta$ projection absorbs leakage at the coarse-grained level at order $\delta$, thereby preserving the Lyapunov structure up to $O(\delta)$ errors.

\subsection{Interpretation as a splitting method}

The $V_\delta$-projected reverse diffusion update
\begin{equation}
  p_{t-1} 
    = \Pi_\delta\bigl( p_t R_t \bigr)
\end{equation}
can also be interpreted as a splitting scheme that approximately minimizes the objective
\begin{equation}
  \mathcal{J}(p) = \mathcal{L}_{\mathrm{fit}}(p) + \lambda\, V_\delta(p),
\end{equation}
where $\mathcal{L}_{\mathrm{fit}}$ represents a loss for ``data fitting'' in the reverse diffusion (e.g., a score-matching loss or an approximation to the log likelihood), and $\lambda$ is a regularization strength.

More specifically,
\begin{enumerate}
  \item the mapping $p_t \mapsto \tilde{p}_{t-1} = p_t R_t$ corresponds to a step that primarily reduces $\mathcal{L}_{\mathrm{fit}}$, and
  \item the projection $\tilde{p}_{t-1} \mapsto p_{t-1} = \Pi_\delta(\tilde{p}_{t-1})$ can be regarded as a Bregman-type projection that reduces $V_\delta$.
\end{enumerate}
In this sense, the proposed method can be viewed as an information-theoretic scheme that balances ``data fitting'' and ``coarse-grained constraints'' in the reverse diffusion of a generative model.

\subsection{Numerical experiments: reproducing image formation in a block model}
\label{subsec:numerics-overview-en}

In this subsection, I verify, using a simple block image model, that the proposed $V_\delta$-projected reverse diffusion can reproduce the ``formation'' of images while controlling coarse-grained quantities. Following the previous work \cite{yourname2025kld}, I start from a model in which an image is partitioned into spatial blocks and:
\begin{itemize}
  \item[(i)] define a forward replication dynamics that converges in the limit to a block-constant configuration;
  \item[(ii)] construct a reverse process starting from random initial conditions, along which block structure emerges.
\end{itemize}
I then compare the behavior with and without the $V_\delta$ projection.

\subsubsection{Block-constant image model}

Consider an image on an $L \times L$ grid, which we partition into $B_x \times B_y$ rectangular blocks. For simplicity we assume that $L$ is divisible by both $B_x$ and $B_y$. Assign a constant intensity $c_j \in [0,1]$ to each block $j$, and set
\begin{equation}
  x_{i,k} = c_j
\end{equation}
for all pixels $(i,k)$ belonging to block $X_j$, thereby generating a block-constant image. The values $c_j$ may be sampled independently from a uniform distribution $\mathrm{Unif}[0,1]$, or from a Bernoulli distribution on $\{0,1\}$, among other choices. In this model, the coarse-grained quantity for block $j$ is simply
\begin{equation}
  w_j(q_{\mathrm{data}}) = \frac{1}{|X_j|} \sum_{(i,k)\in X_j} x_{i,k},
\end{equation}
so that the block average intensity directly corresponds to the block-mass vector of the data distribution.

\subsubsection{Forward replication dynamics and limiting distribution}

Following \cite{yourname2025kld}, I use blockwise Gaussian blurring as the forward replication dynamics. At each step $n$, for an image $x^{(n)}$, a Gaussian kernel $K_j$ is applied only within each block $j$:
\begin{equation}
  x^{(n+1)}_{i,k}
   = \sum_{(u,v)\in X_j} K_j\bigl((i,k),(u,v)\bigr)\, x^{(n)}_{u,v},
   \quad (i,k)\in X_j.
\end{equation}
The kernel $K_j$ is normalized within the block, so that the blockwise mean is preserved at each step. When this dynamics is iterated many times, pixel values become nearly uniform within each block, and the distribution converges to a block-constant limiting state specified only by the block averages.

In the numerical experiments, I consider two types of initial conditions:
\begin{enumerate}
  \item the block-constant image itself as the initial condition, and observe the ``randomization'' under the forward dynamics;
  \item a completely random image (each pixel sampled independently from $\mathrm{Unif}[0,1]$) as the initial condition, and observe the convergence towards a block-constant configuration.
\end{enumerate}
In both cases, the block averages $w_j(p_n)$ are preserved exactly, while the within-block variance decreases and $V(p_n)$ decreases monotonically. This provides a concrete example of the Lyapunov property proved in the previous work, and visualizing it illustrates convergence towards a coarse-grained image layout.

\subsubsection{Reverse kernel and baseline sampler}

Next, I regard the forward blockwise Gaussian blurring as a ``noise-adding process,'' and construct a simple reverse kernel that approximately inverts it. Rather than using deep neural networks, I use a linear approximation to the inverse of the blockwise blur, modeled by blockwise inverse filtering
\begin{equation}
  \tilde{x}^{(n)} = R_n x^{(n+1)},
\end{equation}
with $R_n$ a fixed linear operator at each step. Since $R_n$ is not the exact inverse, this reverse update can produce leakage across block boundaries. In probabilistic terms, one can regard $R_n$ as a linear Gaussian kernel with a small random perturbation,
\begin{equation}
  \tilde{p}_{n} = p_{n+1} R_n,
\end{equation}
which mimics a reverse diffusion with learning errors.

As a baseline sampler, I consider the ``non-projected reverse diffusion,''
\begin{equation}
  p_n^{\mathrm{base}} = \tilde{p}_n,
\end{equation}
which simply takes this $\tilde{p}_n$ as the next state. In the experiments, $p_T$ is taken as the limiting distribution of the forward replication dynamics (nearly block-constant), and the reverse updates are iterated for $n = T-1,\dots,0$ to generate images.

\subsubsection{Comparison with \texorpdfstring{$V_\delta$}{V-delta}-projected reverse diffusion}

In the proposed $V_\delta$-projected reverse diffusion, each step is given by
\begin{equation}
  p_n^{\mathrm{proj}} = \Pi_\delta(\tilde{p}_n),
\end{equation}
where $\tilde{p}_n = p_{n+1}^{\mathrm{proj}} R_n$ is projected onto the leak-tolerant set $\Pi_{P,\delta}(q_{\mathrm{data}})$. In practice, the implementation proceeds by estimating the blockwise mean intensity $w_j(\tilde{p}_n)$ from the pixel samples, computing $w_j^\star$ according to Proposition~\ref{prop:Vdelta-projection-en}, and uniformly scaling the pixel values within each block so that the new block masses satisfy $w_j\bigl(\Pi_\delta(\tilde{p}_n)\bigr) = w_j^\star$.

I compare the following two processes:
\begin{itemize}
  \item the non-projected reverse diffusion $p_n^{\mathrm{base}}$ (baseline);
  \item the $V_\delta$-projected reverse diffusion $p_n^{\mathrm{proj}}$.
\end{itemize}
The comparison is based on the following indicators:
\begin{enumerate}
  \item the block-mass error
  \begin{equation}
    E_{\mathrm{block}}(n) 
      = \bigl\| w\bigl(p_n\bigr) - w(q_{\mathrm{data}}) \bigr\|_1,
  \end{equation}
  \item the evolution of the potentials $V(p_n)$ and $V_\delta(p_n)$,
  \item the pixel-level mean squared error
  \begin{equation}
    E_{\mathrm{pix}}(n)
      = \frac{1}{L^2} \sum_{i,k}
        \bigl(x_{i,k}^{(n)} - x_{i,k}^{\mathrm{data}}\bigr)^2,
  \end{equation}
\end{enumerate}
where $x^{\mathrm{data}}$ denotes the original block-constant image.

We expect that, in the baseline process $p_n^{\mathrm{base}}$, block means gradually deteriorate as the reverse process progresses, and $E_{\mathrm{block}}(n)$ and $V_\delta(p_n)$ can increase. In contrast, for $p_n^{\mathrm{proj}}$, $E_{\mathrm{block}}(n)$ should remain of order $\delta$, $V_\delta(p_n)$ should be almost monotone, and $V(p_n)$ should decrease up to $O(\delta)$ errors. As for $E_{\mathrm{pix}}(n)$ and visual image quality, we expect both processes to be comparable, with a possibility that the projected process yields slightly better results by preventing deterioration of local contrast.

\subsubsection{Image formation from random initial conditions}

By combining the forward replication dynamics and reverse diffusion, one can construct a process that exhibits ``image formation with block structure from random initial states.'' Concretely, consider the following steps:
\begin{enumerate}
  \item Sample a random initial image $x^{(0)}$, with each pixel drawn independently from a uniform or Gaussian distribution.
  \item Apply the forward replication dynamics for $T$ steps to obtain $x^{(T)}$, where $T$ is chosen large enough that the configuration is nearly block-constant.
  \item Use $x^{(T)}$ as the initial condition and run the non-projected reverse diffusion $p_n^{\mathrm{base}}$ and the $V_\delta$-projected reverse diffusion $p_n^{\mathrm{proj}}$ separately.
\end{enumerate}
This setup constructs a round trip process in which random noise is first mapped to a nearly block-constant configuration, and then finer structures are reintroduced by reverse diffusion.

In the numerical experiments, I visualize
\begin{itemize}
  \item the random initial image and the final generated images;
  \item the time evolution of block masses and the potentials $V$ and $V_\delta$;
  \item the relaxation and reconstruction of within-block variances and local histograms,
\end{itemize}
to qualitatively reproduce an ``image formation process'' from the viewpoint of coarse-grained quantities. For visualization, I plot $V(p_n)$ and $E_{\mathrm{block}}(n)$ as functions of $n$, and display snapshots of images at representative steps. The effect of the $V_\delta$ projection can be intuitively understood by comparing these plots.

\subsection{Model quantities}
\label{subsec:quantities-en}

In this subsection, I define the quantities
\(
V(p),\ V_\delta(p),\ E_{\mathrm{block}}(n)
\)
that are evaluated numerically in the block image model. These are based on the KLD potential framework introduced in \cite{yourname2025kld}, and are used here as indicators of stability for the reverse diffusion.

\subsubsection{Block partition and block masses}

We regard an image as a set of pixels on an \(L \times L\) grid, partitioned into \(B_x \times B_y\) rectangular blocks. That is, the state space
\[
  X = \bigsqcup_{j=1}^{m} X_j,
  \qquad m = B_x B_y
\]
is defined as a disjoint union of blocks \(X_j\).

A normalized image is treated as a probability vector
\(
p \in \Delta(X)
\)
with components
\(
p_{i,k} \ge 0
\)
for each pixel \((i,k)\). The normalization condition is
\[
  \sum_{(i,k)\in X} p_{i,k} = 1.
\]

The probability mass in block \(j\) (the coarse-grained quantity) is defined by
\begin{equation}
  w_j(p)
  = \sum_{(i,k) \in X_j} p_{i,k},
  \qquad j = 1,\dots,m,
  \label{eq:block-mass-en}
\end{equation}
and the vector
\(
w(p) = (w_1(p),\dots,w_m(p))
\)
is called the block-mass vector.

For the data distribution \(q_{\mathrm{data}}\), the block masses are
\begin{equation}
  w_j^{\mathrm{ref}}
  = w_j\bigl(q_{\mathrm{data}}\bigr),
  \qquad j = 1,\dots,m,
  \label{eq:wref-def-en}
\end{equation}
which are regarded as the ``reference coarse-grained quantities.'' In the block image experiments, \(q_{\mathrm{data}}\) is constructed as a block-constant image, and \(w_j^{\mathrm{ref}}\) is computed from the block averages.

\subsubsection{\texorpdfstring{$V(p)$}{V(p)}: potential for within-block non-uniformity}

In the non-ergodic image model of \cite{yourname2025kld}, a potential \(V(p)\) was defined based on the KLD between the within-block conditional distribution and the uniform distribution. Here I adopt the same definition. For block \(b\),
\begin{align}
  w_b &= w_b(p_n)
       = \sum_{x \in X_b} p_n(x), \label{eq:block-mass-b-en} \\
  p_b(x) &= \frac{p_n(x)}{w_b}, \quad x \in X_b, \label{eq:block-conditional-en} \\
  u_b(x) &= \frac{1}{|X_b|}, \quad x \in X_b, \label{eq:block-uniform-en}
\end{align}
where \(p_n\) is the distribution at time \(n\), \(p_b\) is the conditional histogram within block $b$, and \(u_b\) is the uniform distribution on the same block.

Using the KLD
\(D_{\mathrm{KL}}(p\Vert q)\), define
\begin{equation}
  V(p_n)
  = \sum_b w_b\,
      D_{\mathrm{KL}}\bigl(p_b \,\Vert\, u_b\bigr).
  \label{eq:V-block-uniform-en}
\end{equation}
This potential measures the degree of non-uniformity within blocks under fixed block masses (no leakage). In the forward blockwise Gaussian blurring, $V(p_n)$ decreases monotonically, as shown in \cite{yourname2025kld}.

\subsubsection{\texorpdfstring{$V_\delta(p)$}{V\_delta(p)}: leak-tolerant coarse-grained potential}

In practical reverse diffusion, leakage across block boundaries occurs due to learning and discretization errors, so that the block masses \(w_j(p_n)\) are not exactly conserved. To handle this situation, \cite{yourname2025kld} introduced a leak-tolerant set
\begin{equation}
  \Pi_{P,\delta}(p_0)
  = \left\{
      \pi \in \Delta(X) :
      \forall j,\ 
      \bigl|w_j(\pi) - w_j(p_0)\bigr| \le \delta
    \right\},
  \label{eq:PiPdelta-en}
\end{equation}
and defined the leak-tolerant potential as
\begin{equation}
  V_\delta(p)
  = \inf_{\pi \in \Pi_{P,\delta}(p_0)}
      D_{\mathrm{KL}}(p \Vert \pi).
  \label{eq:Vdelta-def-en}
\end{equation}

At the coarse-grained level, the potential admits a closed-form expression in terms of block masses \cite[Eq.~(19)]{yourname2025kld}. Let \(w_j(p)\) denote the block masses of \(p\), and define
\begin{equation}
  a_j = w_j^{\mathrm{ref}} - \delta,
  \qquad
  b_j = w_j^{\mathrm{ref}} + \delta,
  \label{eq:aj-bj-def-en}
\end{equation}
for the reference masses \(w_j^{\mathrm{ref}}\). For a positive scalar \(\tau > 0\),
\begin{equation}
  \hat{w}_j(\tau)
   = \biggl[ \frac{w_j(p)}{\tau} \biggr]_{[a_j,b_j]}
   = \min\bigl\{ \max\{ w_j(p)/\tau,\ a_j\},\ b_j\bigr\},
  \label{eq:w-hat-en}
\end{equation}
and the equation
\begin{equation}
  \sum_{j=1}^{m} \hat{w}_j(\tau) = 1
  \label{eq:tau-star-eq-en}
\end{equation}
has a unique solution \(\tau^\star > 0\). The optimal block masses are then
\begin{equation}
  w_j^\star
  = \hat{w}_j(\tau^\star),
  \qquad j = 1,\dots,m.
  \label{eq:w-star-def-en}
\end{equation}
The coarse-grained part of the potential can be written as
\begin{equation}
  V_\delta(p)
  = \sum_{j=1}^{m}
      w_j(p)\,
      \log \frac{w_j(p)}{w_j^\star}.
  \label{eq:Vdelta-coarse-en}
\end{equation}

In the present experiments, the reference masses \(w_j^{\mathrm{ref}}\) are computed from the original block-constant image as in \eqref{eq:wref-def-en}, and the tolerance \(\delta\) is fixed to a small value (e.g., \(\delta = 0.01\)).

\subsubsection{Block error \texorpdfstring{$E_{\mathrm{block}}(n)$}{E\_block(n)}}

To directly quantify the leakage of block masses at each reverse step \(n\), define
\begin{equation}
  E_{\mathrm{block}}(n)
  = \bigl\|
      w\bigl(p_n\bigr) - w^{\mathrm{ref}}
    \bigr\|_1
  = \sum_{j=1}^{m}
      \bigl|w_j(p_n) - w_j^{\mathrm{ref}}\bigr|,
  \label{eq:Eblock-def-en}
\end{equation}
where \(\|\cdot\|_1\) denotes the $L^1$ norm and \(w(p_n)\) is the block-mass vector at time \(n\).

The quantity \(E_{\mathrm{block}}(n)\) measures ``how much the block masses have leaked from the reference structure,'' and comparing it with the tolerance \(\delta\) allows us to check whether
\begin{equation}
  E_{\mathrm{block}}(n) \lesssim m\,\delta
\end{equation}
holds. In the numerical experiments, I plot the time evolution of \(V(p_n)\), \(V_\delta(p_n)\), and \(E_{\mathrm{block}}(n)\) and compare the behavior of the non-projected and $V_\delta$-projected reverse diffusion.

\subsection{Algorithm for reverse diffusion from block-constant images}
\label{subsec:algo-reverse-snapshots-en}

\begin{figure}
 \centering
  % Left: data image q_data, middle: initial state p_0, right: reverse-diffusion snapshots
  \includegraphics[width=\textwidth]{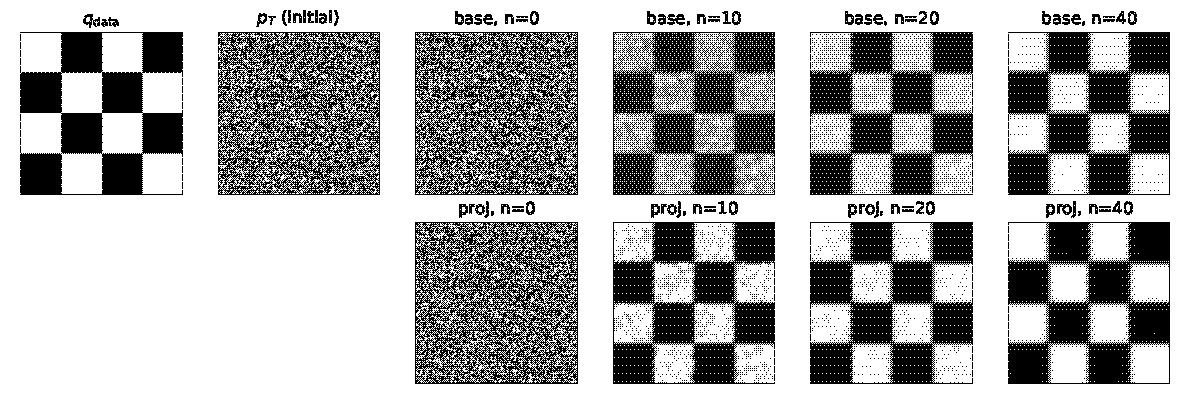}
  \caption{
    Example of reverse diffusion from a block-constant image.
    Left: original block-constant image $q_{\mathrm{data}}$.
    Middle: initial state $p_0$ obtained by blockwise Gaussian blurring.
    Right: snapshots at representative steps $n$ for the non-projected reverse diffusion and the $V_\delta$-projected reverse diffusion.
  }
  \label{fig:reverse-snapshots}
\end{figure}

The numerical experiment ``reverse diffusion from a block-constant image,'' shown schematically in Fig.~\ref{fig:reverse-snapshots}, is constructed as follows.

\paragraph{Step 1: Generation of the data image $q_{\mathrm{data}}$ (left panel)}

\begin{enumerate}
  \item Choose the image size $L \times L$ and the block partition $B_x \times B_y$.
        Define the state space as
        \(
          X = \bigsqcup_{j=1}^{m} X_j,\ m = B_x B_y,
        \)
        where $X_j$ is the set of pixels in block $j$.
  \item For each block $j$, assign a constant intensity $c_j$. For example,
        \begin{itemize}
          \item arrange $c_j \in \{c_{\mathrm{high}}, c_{\mathrm{low}}\}$ in a checkerboard pattern;
          \item or sample $c_j \sim \mathrm{Unif}[0,1]$ independently.
        \end{itemize}
  \item Set
        \(
          x_{i,k} = c_j
        \)
        for all pixels $(i,k)\in X_j$ to obtain a block-constant image, which is taken as the data image $q_{\mathrm{data}}$.
  \item If necessary, normalize $q_{\mathrm{data}}$ to a probability distribution by
        \(
          q_{\mathrm{data}}(i,k) = x_{i,k} / \sum_{(u,v)} x_{u,v}.
        \)
  \item Compute the reference block masses
        \(
          w_j^{\mathrm{ref}} = w_j(q_{\mathrm{data}})
        \)
        according to \eqref{eq:block-mass-en} and \eqref{eq:wref-def-en}, and store them as the coarse-grained quantities used in the experiments.
\end{enumerate}

\paragraph{Step 2: Generation of the initial state $p_0$ (middle panel)}

\begin{enumerate}
  \item Starting from $q_{\mathrm{data}}$, apply a forward replication step given by blockwise Gaussian blurring. For each block $X_j$,
        \begin{equation}
          p_0(i,k)
          = \sum_{(u,v)\in X_j}
              K_j\bigl((i,k),(u,v)\bigr)\,
              q_{\mathrm{data}}(u,v),
          \qquad (i,k)\in X_j,
        \end{equation}
        where $K_j$ is a Gaussian kernel normalized within block $j$.
  \item If desired, apply multiple steps of the blockwise Gaussian to obtain a more randomized state $p_0$.
  \item Normalize $p_0$ so that
        \(
          \sum_{(i,k)} p_0(i,k) = 1.
        \)
        This $p_0$ is used as the initial state for the reverse diffusion (middle panel of Fig.~\ref{fig:reverse-snapshots}).
\end{enumerate}

\paragraph{Step 3: Non-projected and $V_\delta$-projected reverse diffusion (right panels)}

\begin{enumerate}
  \item Choose the number of reverse steps $T$ and define $t = T,T-1,\dots,0$.
        Set the initial conditions
        \(
          p^{\mathrm{base}}_T = p^{\mathrm{proj}}_T = p_0.
        \)
  \item For each step $t = T,T-1,\dots,1$, perform:
        \begin{enumerate}
          \item \textbf{Non-projected reverse diffusion (baseline):}
                For the current distribution $p^{\mathrm{base}}_t$, apply a simplified reverse kernel $R_t$,
                \begin{equation}
                  p^{\mathrm{base}}_{t-1}
                  = p^{\mathrm{base}}_{t} R_t.
                \end{equation}
                As a simple model, we use an unsharp-masking-like step that approximately inverts the forward Gaussian blur,
                \[
                  \tilde{p} = p - \alpha\,(G_\sigma p - p),
                \]
                as a concrete form of $R_t$.
          \item \textbf{$V_\delta$-projected reverse diffusion:}
                Using the same kernel $R_t$, first compute
                \begin{equation}
                  \tilde{p}_{t-1}
                  = p^{\mathrm{proj}}_t R_t,
                \end{equation}
                and then apply the KLD projection based on the leak-tolerant potential $V_\delta$,
                \begin{equation}
                  p^{\mathrm{proj}}_{t-1}
                  = \Pi_\delta\bigl(\tilde{p}_{t-1}\bigr).
                \end{equation}
                Here $\Pi_\delta$ is defined as the information projection onto
                \(\Pi_{P,\delta}(q_{\mathrm{data}})\):
                \begin{equation}
                  \Pi_\delta(p)
                  = \arg\min_{\pi \in \Pi_{P,\delta}(q_{\mathrm{data}})}
                      D_{\mathrm{KL}}(p \Vert \pi).
                \end{equation}
                In implementation, the block masses $w_j(p)$ are computed, the optimal masses \(w_j^\star\) are obtained via \eqref{eq:aj-bj-def-en}--\eqref{eq:w-star-def-en}, and pixel values within each block are scaled uniformly so that
                \(
                  w_j\bigl(\Pi_\delta(p)\bigr) = w_j^\star.
                \)
        \end{enumerate}
  \item At a few representative steps
        \(
          t = T,\, t_1,\, t_2,\, \dots,\, 0,
        \)
        save images of
        \begin{itemize}
          \item the non-projected reverse diffusion $p^{\mathrm{base}}_t$,
          \item the $V_\delta$-projected reverse diffusion $p^{\mathrm{proj}}_t$,
        \end{itemize}
        and display them in Fig.~\ref{fig:reverse-snapshots}. Placing $q_{\mathrm{data}}$ at the left and $p_0$ in the middle, we can visually inspect the process
        \[
          q_{\mathrm{data}}
            \ \longrightarrow\ 
          p_0
            \ \longrightarrow\ 
          p_t^{\mathrm{base}},\ p_t^{\mathrm{proj}},
        \]
        i.e., reverse diffusion from a block-constant image.
\end{enumerate}

\begin{figure}
    \centering
    \includegraphics[width=0.5\linewidth]{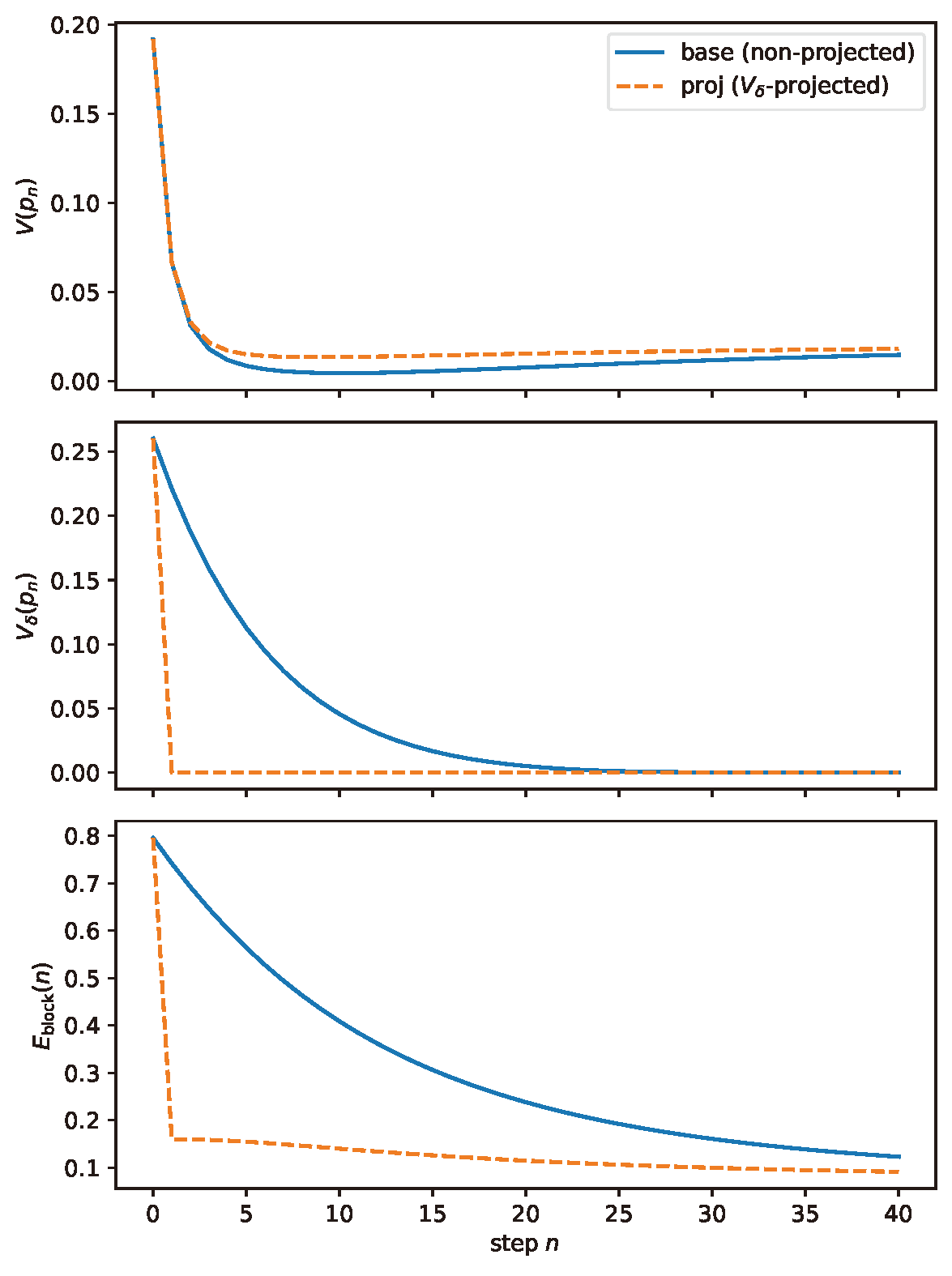}
    \caption{
    Time evolution of information-theoretic indicators during reverse diffusion.
    Top: $V(p_n)$.
    Middle: leak-tolerant potential $V_\delta(p_n)$.
    Bottom: block-mass error
    $E_{\mathrm{block}}(n) = \|w(p_n) - w^{\mathrm{ref}}\|_1$.
    Solid lines correspond to the non-projected reverse diffusion; dashed lines correspond to the $V_\delta$-projected reverse diffusion.
    In the projected case, both $E_{\mathrm{block}}(n)$ and $V_\delta(p_n)$ remain within the tolerance controlled by $\delta$, yielding a stable generative process from the viewpoint of coarse-grained quantities.
  }
    \label{fig:placeholder}
\end{figure}

\section{Discussion and future work}
\label{sec:discussion}

In this study, I attempted to transplant the framework of non-ergodic replication dynamics and the KLD potentials $V$ and $V_\delta$ introduced in \cite{yourname2025kld} to the reverse diffusion of generative models. Specifically, I presented (i) an extension of the Lyapunov property of $V$ to time-dependent block-preserving kernels, (ii) a closed-form expression for the KLD projection $\Pi_\delta$ based on the leak-tolerant potential $V_\delta$, and (iii) the design of a $V_\delta$-projected reverse diffusion scheme in which $\Pi_\delta$ is inserted after each reverse step. In numerical experiments on a block-constant image model, I confirmed that $V_\delta$ projection keeps the block-mass error $E_{\mathrm{block}}(n)$ within the tolerance $\delta$ and maintains $V_\delta(p_n)$ almost monotone, supporting the main aim of this work: stabilizing the generative process from the viewpoint of coarse-grained quantities.

The theoretical contributions can be summarized in two points. First, Theorem~\ref{thm:time-inhomogeneous-en} extends the previous time-homogeneous setting to time-dependent block-preserving kernels $T_n$, showing that $V$ can act as a Lyapunov function even for time-inhomogeneous controlled Markov processes such as reverse diffusion. Second, Theorem~\ref{thm:Vdelta-monotone-en} shows that, even in the presence of leakage, the pseudo-monotonicity of $V_\delta$ is preserved as long as deviations of block masses are controlled at order $\delta$. This formalizes the intuition that the $V_\delta$ projection plays the role of ``absorbing leakage at the coarse-grained level,'' thereby maintaining the original Lyapunov structure up to $O(\delta)$ errors.

In the numerical experiments, instead of using actual DDPM \cite{ho2020ddpm,nichol2021improved} or score-based models \cite{song2021scorebased}, I adopted a toy reverse step based on the geometric mean of the current distribution $p_t$ and the data distribution $q_{\mathrm{data}}$,
\(
p_{t-1} \propto p_t^{1-\beta} q_{\mathrm{data}}^{\beta},
\)
combined with a weak Gaussian smoothing. This choice is a simplified discrete analogue of the flows that decrease the KL divergence to a reference distribution seen in Schr{\"o}dinger-bridge-type generative modeling \cite{debortoli2021dsb,wang2021dglsb}, and, in the conditional setting, is also related to flows studied in the context of conditional sampling \cite{zhao2024conditional}. By placing the $V_\delta$-based KLD projection after this step, the reverse process is constructed as a splitting method that alternately satisfies ``data fitting'' and ``coarse-grained constraints.'' The experiments show that this scheme is consistent with the coarse-grained Lyapunov structure, and can stably reproduce the emergence of block structure from random initial states.

There are several limitations to the present work. First, both the theory and experiments are restricted to block image models on finite state spaces, and do not directly treat continuous-space score-based SDEs \cite{song2021scorebased} or large-scale image datasets. In a continuous-time setting, one would need to describe the time evolution of $V$ and $V_\delta$ at the level of PDEs corresponding to the Fokker--Planck equation, and connect the analysis to the theory of reaction--diffusion systems and asymptotic stability. Second, the reverse step considered here is a hand-crafted linear or quasi-linear kernel and differs greatly from realistic neural-network-based score approximations. In practical diffusion models, network approximation errors and sampling discretization errors appear in a nonlinear and high-dimensional manner. Quantitatively assessing how well these effects can be approximated as block-level leakage of probability mass remains an important problem for future work.

As future directions, I highlight at least three lines of research. First, there is the design of coarse-grained quantities in realistic DDPM and score-based models. In this paper, I used simple spatial block averages, but for real images one can define higher-level coarse-grained quantities such as segment-wise class proportions, low-frequency histograms after filtering, or feature maps from pretrained networks. These quantities naturally relate to existing concepts in the general framework of conditional diffusion \cite{zhao2024conditional}, histogram-constrained image generation \cite{hig2026}, and manifold-constrained classifier-free guidance \cite{chung2025cfgpp}. Applying $V_\delta$ projection to these coarse-grained quantities may enable generative models to ``sample fine details while preserving global layouts and class compositions.''

Second, one can extend the framework to \emph{learn} the coarse-graining itself. For example, autoencoders or self-supervised representation learning can be used to extract low-dimensional macroscopic variables from images, and $V_\delta$ can be defined in this learned space and incorporated into reverse diffusion, thus producing a coarse-grained potential driven by data.

Third, the connection to Schr{\"o}dinger bridges and optimal transport with entropy regularization deserves further investigation. In these frameworks, the path-space KL divergence between a reference process and a target distribution is minimized, while the $V_\delta$ proposed here is characterized by using the coarse-grained KL of intermediate distributions as a Lyapunov function. Bridging these perspectives could lead to a new class of generative schemes such as ``coarse-grained Schr{\"o}dinger bridges.'' Furthermore, combining this approach with Lyapunov-based diffusion control \cite{liu2025lyapdiff} may pave the way to sampler designs that guarantee both stability of coarse-grained quantities and control-theoretic safety.

Finally, the information-theoretic indicators introduced in this paper, $V$, $V_\delta$, and $E_{\mathrm{block}}(n)$, are expected to be useful not only as Lyapunov functions for sampler design, but also as evaluation metrics for generative models. Conventional metrics such as Fr\'echet Inception Distance and Inception Score \cite{heusel2017fid,salimans2016is} are mainly based on the mean and covariance in a feature space, whereas the potentials considered here are defined directly from the coarse-grained structures inherent in the data. In future work, by systematically extracting coarse-grained statistics from real image datasets and comparing the behavior of $V_\delta$ across different models, one may evaluate generative models from a new viewpoint: ``how faithful they are to the data at the level of coarse-grained physical quantities.''

\bibliographystyle{plain}
\bibliography{references}

\section*{Appendix}

Code for Appendix Figure~1:
\begin{verbatim}
import numpy as np
from scipy.ndimage import gaussian_filter
import matplotlib.pyplot as plt

def normalize_pmf(img):
    """Normalize a nonnegative image as a probability distribution (sum = 1)."""
    img = np.clip(img, 1e-12, None)
    s = img.sum()
    if s <= 0:
        raise ValueError("sum <= 0 in normalize_pmf")
    return img / s


def make_blocks(Lx, Ly, Bx, By):
    """
    Return the list of (x-slice, y-slice) pairs when an Lx x Ly image
    is partitioned into Bx x By blocks.
    """
    assert Lx % Bx == 0 and Ly % By == 0
    hx, hy = Lx // Bx, Ly // By
    blocks = []
    for bx in range(Bx):
        for by in range(By):
            xs = slice(bx * hx, (bx + 1) * hx)
            ys = slice(by * hy, (by + 1) * hy)
            blocks.append((xs, ys))
    return blocks


def block_masses(p, blocks):
    """Return the block mass vector w(p)."""
    w = []
    for xs, ys in blocks:
        w.append(p[xs, ys].sum())
    return np.array(w)


# ============================================
# Generation of block-constant (checkerboard) images
# ============================================

def init_block_constant_image(Lx=128, Ly=128, Bx=4, By=4,
                              mode="checkerboard", seed=7):
    """
    Generate an image that is constant or binary-valued on each block.
    mode="checkerboard": place high-/low-intensity values alternately on even/odd blocks.
    mode="constant":     assign a constant drawn from a uniform random variable on each block.
    """
    rng = np.random.default_rng(seed)
    blocks = make_blocks(Lx, Ly, Bx, By)
    img = np.zeros((Lx, Ly), dtype=float)

    for bi, (xs, ys) in enumerate(blocks):
        bx = bi // By
        by = bi % By
        if mode == "checkerboard":
            # Checkerboard: brighten blocks with even (bx + by)
            val = 0.9 if (bx + by) % 2 == 0 else 0.1
            block = np.full((xs.stop - xs.start,
                             ys.stop - ys.start), val)
        elif mode == "constant":
            c = rng.uniform(0.0, 1.0)
            block = np.full((xs.stop - xs.start,
                             ys.stop - ys.start), c)
        else:
            raise ValueError("unknown mode")
        img[xs, ys] = block

    return img, blocks


# ============================================
# V_delta projection (rescale + clip of block masses)
# ============================================

def project_block_masses(p, blocks, w_ref, delta):
    """
    V_delta projection Pi_delta(p) that scales only the block masses,
    keeping the within-block shape fixed, using the optimal block masses w*_j of V_delta.
    Corresponds to equation (19) of yourname2025kld.
    """
    w = block_masses(p, blocks)
    # Allowable interval [a_j, b_j]
    a = np.clip(w_ref - delta, 0.0, 1.0)
    b = np.clip(w_ref + delta, 0.0, 1.0)

    def f(tau):
        w_over = w / tau
        w_star = np.minimum(np.maximum(w_over, a), b)
        return w_star.sum() - 1.0

    # Coarsely bracket tau and solve by bisection
    tau_lo, tau_hi = 1e-6, 1e6

    if f(tau_lo) < 0:
        while f(tau_lo) < 0 and tau_lo < 1e3:
            tau_lo *= 10.0
    if f(tau_hi) > 0:
        while f(tau_hi) > 0 and tau_hi > 1e-6:
            tau_hi *= 0.1

    for _ in range(80):
        tau_mid = 0.5 * (tau_lo + tau_hi)
        val = f(tau_mid)
        if abs(val) < 1e-10:
            break
        if val > 0:
            tau_lo = tau_mid
        else:
            tau_hi = tau_mid

    tau_star = tau_mid
    w_over = w / tau_star
    w_star = np.minimum(np.maximum(w_over, a), b)

    # Scale in each block
    p_new = np.zeros_like(p)
    for idx, (xs, ys) in enumerate(blocks):
        block = p[xs, ys]
        w_b = w[idx]
        if w_b <= 0:
            continue
        scale = w_star[idx] / w_b
        p_new[xs, ys] = block * scale

    p_new = normalize_pmf(p_new)
    return p_new


# ============================================
# Reverse step that moves toward the data
# ============================================

def reverse_step_toward_data(p, q_data, beta=0.05, sigma_smooth=0.5):
    """
    Reverse step that ``moves a coarse state p slightly toward the data q_data''.

    Based on the idealized discrete update
        p_{t-1}(x) ~ p_t(x)^{1-beta} q_data(x)^{beta},
    with an additional mild Gaussian smoothing.
    """
    # Move toward q_data via geometric mean
    logp = np.log(p)
    logq = np.log(q_data)
    mix_log = (1.0 - beta) * logp + beta * logq
    mixed = np.exp(mix_log)
    mixed = normalize_pmf(mixed)

    # Apply mild smoothing (to even out noise)
    if sigma_smooth > 0:
        smoothed = gaussian_filter(mixed, sigma=sigma_smooth, mode="reflect")
        smoothed = normalize_pmf(smoothed)
        return smoothed
    else:
        return mixed


# ============================================
# Generate base / proj trajectories from a coarse initial state
# ============================================

def run_reverse_paths_restore(
    q_data, blocks, w_ref,
    nsteps=40, delta=0.01,
    beta=0.05, sigma_smooth=0.5,
    seed=0
):
    """
    Run reverse diffusion that ``moves toward the data'' from a rough initial state,
    given q_data (clean checker), its block partition blocks, and w_ref.
      base: moves toward the data only
      proj: moves toward the data + V_delta projection
    """
    rng = np.random.default_rng(seed)
    Lx, Ly = q_data.shape

    # --- Construct a rough initial state p_T ---
    # Example: pure noise
    noise = rng.random((Lx, Ly))
    p_T = normalize_pmf(noise)

    # Store the base/proj trajectories
    path_base = np.zeros((nsteps + 1, Lx, Ly))
    path_proj = np.zeros((nsteps + 1, Lx, Ly))

    p_base = p_T.copy()
    p_proj = p_T.copy()

    path_base[0] = p_base
    path_proj[0] = p_proj

    for n in range(1, nsteps + 1):
        # base: move toward the data only
        p_base = reverse_step_toward_data(
            p_base, q_data, beta=beta, sigma_smooth=sigma_smooth
        )

        # proj: move toward the data + V_delta projection
        tmp = reverse_step_toward_data(
            p_proj, q_data, beta=beta, sigma_smooth=sigma_smooth
        )
        p_proj = project_block_masses(tmp, blocks, w_ref, delta)

        path_base[n] = p_base
        path_proj[n] = p_proj

    return p_T, path_base, path_proj


# ============================================
# Plot snapshot figure
# ============================================

def plot_reverse_snapshots_restore(
    q_data, p_T, path_base, path_proj,
    snapshot_steps=(0, 10, 20, 40),
    fname="reverse_snapshots_restore.pdf"
):
    """
    Left:  q_data (clean checkerboard)
    Center: p_T (rough initial state)
    Right: snapshots of base/proj.
    """
    Lx, Ly = q_data.shape
    ncols = 2 + len(snapshot_steps)
    fig, axes = plt.subplots(2, ncols, figsize=(2.0 * ncols, 4.0))

    for ax_row in axes:
        for ax in ax_row:
            ax.set_xticks([])
            ax.set_yticks([])

    vmin, vmax = q_data.min(), q_data.max()

    # Left: data image
    axes[0, 0].imshow(q_data, cmap="gray", vmin=vmin, vmax=vmax)
    axes[0, 0].set_title(r"$q_{\mathrm{data}}$")
    axes[1, 0].axis("off")

    # Center: initial state p_T
    axes[0, 1].imshow(p_T, cmap="gray", vmin=vmin, vmax=vmax)
    axes[0, 1].set_title(r"$p_T$ (initial)")
    axes[1, 1].axis("off")

    # Right: snapshots
    for j, step in enumerate(snapshot_steps):
        col = 2 + j
        step = min(step, path_base.shape[0] - 1)

        # Top row: base
        axes[0, col].imshow(path_base[step], cmap="gray",
                            vmin=vmin, vmax=vmax)
        axes[0, col].set_title(f"base, n={step}")

        # Bottom row: proj
        axes[1, col].imshow(path_proj[step], cmap="gray",
                            vmin=vmin, vmax=vmax)
        axes[1, col].set_title(f"proj, n={step}")

    plt.tight_layout()
    plt.savefig(fname, bbox_inches="tight")
    plt.close(fig)


# ============================================
# Main
# ============================================

if __name__ == "__main__":
    # Image size and block partition
    Lx = Ly = 128
    Bx = By = 4

    # Generate a clean checkerboard image q_data
    img_data, blocks = init_block_constant_image(
        Lx=Lx, Ly=Ly, Bx=Bx, By=By,
        mode="checkerboard", seed=7
    )
    q_data = normalize_pmf(img_data)
    w_ref = block_masses(q_data, blocks)

    # Run reverse diffusion from a rough initial state (base / proj)
    p_T, path_base, path_proj = run_reverse_paths_restore(
        q_data, blocks, w_ref,
        nsteps=40, delta=0.01,
        beta=0.05, sigma_smooth=0.5,
        seed=0
    )

    # Output snapshot figure
    plot_reverse_snapshots_restore(
        q_data, p_T, path_base, path_proj,
        snapshot_steps=(0, 10, 20, 40),
        fname="reverse_snapshots_restore.pdf"
    )
    print("Saved reverse_snapshots_restore.pdf")
\end{verbatim}

Appendix: code for Figure~2
\begin{verbatim}
import numpy as np
from scipy.ndimage import gaussian_filter
import matplotlib.pyplot as plt

# ============================================
# Basic utilities
# ============================================

def normalize_pmf(img):
    """Normalize a nonnegative image as a probability distribution (sum = 1)."""
    img = np.clip(img, 1e-12, None)
    s = img.sum()
    if s <= 0:
        raise ValueError("sum <= 0 in normalize_pmf")
    return img / s


def make_blocks(Lx, Ly, Bx, By):
    """
    Return the list of (x-slice, y-slice) pairs when an Lx x Ly image
    is partitioned into Bx x By blocks.
    """
    assert Lx % Bx == 0 and Ly % By == 0
    hx, hy = Lx // Bx, Ly // By
    blocks = []
    for bx in range(Bx):
        for by in range(By):
            xs = slice(bx * hx, (bx + 1) * hx)
            ys = slice(by * hy, (by + 1) * hy)
            blocks.append((xs, ys))
    return blocks


def block_masses(p, blocks):
    """Return the block mass vector w(p)."""
    w = []
    for xs, ys in blocks:
        w.append(p[xs, ys].sum())
    return np.array(w)


# ============================================
# Generation of block-constant (checkerboard) images
# ============================================

def init_block_constant_image(Lx=128, Ly=128, Bx=4, By=4,
                              mode="checkerboard", seed=7):
    """
    Generate an image that is constant or binary-valued on each block.
    mode="checkerboard": place high-/low-intensity values alternately on even/odd blocks.
    mode="constant":     assign a constant drawn from a uniform random variable on each block.
    """
    rng = np.random.default_rng(seed)
    blocks = make_blocks(Lx, Ly, Bx, By)
    img = np.zeros((Lx, Ly), dtype=float)

    for bi, (xs, ys) in enumerate(blocks):
        bx = bi // By
        by = bi % By
        if mode == "checkerboard":
            # Checkerboard: brighten blocks with even (bx + by)
            val = 0.9 if (bx + by) % 2 == 0 else 0.1
            block = np.full((xs.stop - xs.start,
                             ys.stop - ys.start), val)
        elif mode == "constant":
            c = rng.uniform(0.0, 1.0)
            block = np.full((xs.stop - xs.start,
                             ys.stop - ys.start), c)
        else:
            raise ValueError("unknown mode")
        img[xs, ys] = block

    return img, blocks


# ============================================
# Definitions of V, V_delta, and E_block
# ============================================

def kl_div(p, q):
    """D_KL(p || q) for flat arrays."""
    p = np.asarray(p, dtype=float).ravel()
    q = np.asarray(q, dtype=float).ravel()
    mask = (p > 0) & (q > 0)
    return np.sum(p[mask] * (np.log(p[mask]) - np.log(q[mask])))


def V_block_uniform(p, blocks):
    """
    V(p) = sum_b w_b D_KL(p_b || u_b)
    Potential based on the KLD to the within-block uniform distribution.
    """
    V = 0.0
    for xs, ys in blocks:
        block = p[xs, ys]
        w_b = block.sum()
        if w_b <= 0:
            continue
        p_b = block / w_b
        u_b = np.full_like(block, 1.0 / block.size)
        V += w_b * kl_div(p_b, u_b)
    return V


def compute_w_star(w, w_ref, delta, tol=1e-10, max_iter=80):
    """
    Compute the optimal block masses w*_j appearing in the closed form of V_delta.
    w:      current block mass vector
    w_ref:  reference block mass vector
    delta:  tolerance width delta
    """
    a = np.clip(w_ref - delta, 0.0, 1.0)
    b = np.clip(w_ref + delta, 0.0, 1.0)

    def f(tau):
        w_over = w / tau
        w_star = np.minimum(np.maximum(w_over, a), b)
        return w_star.sum() - 1.0

    tau_lo, tau_hi = 1e-6, 1e6

    if f(tau_lo) < 0:
        while f(tau_lo) < 0 and tau_lo < 1e3:
            tau_lo *= 10.0
    if f(tau_hi) > 0:
        while f(tau_hi) > 0 and tau_hi > 1e-6:
            tau_hi *= 0.1

    for _ in range(max_iter):
        tau_mid = 0.5 * (tau_lo + tau_hi)
        val = f(tau_mid)
        if abs(val) < tol:
            break
        if val > 0:
            tau_lo = tau_mid
        else:
            tau_hi = tau_mid

    tau_star = tau_mid
    w_over = w / tau_star
    w_star = np.minimum(np.maximum(w_over, a), b)
    return w_star


def V_delta_coarse(p, blocks, w_ref, delta):
    """
    Compute the coarse-grained part of V_delta(p)
    (V_delta = sum_j w_j log(w_j / w*_j)).
    """
    w = block_masses(p, blocks)
    w_star = compute_w_star(w, w_ref, delta)
    mask = (w > 0) & (w_star > 0)
    return np.sum(w[mask] * (np.log(w[mask]) - np.log(w_star[mask])))


def E_block(p, blocks, w_ref):
    """
    E_block = || w(p) - w_ref ||_1
    """
    w = block_masses(p, blocks)
    return np.sum(np.abs(w - w_ref))


# ============================================
# V_delta projection (rescale + clip of block masses)
# ============================================

def project_block_masses(p, blocks, w_ref, delta):
    """
    V_delta projection Pi_delta(p) that scales only the block masses,
    keeping the within-block shape fixed, using the optimal block masses w*_j of V_delta.
    """
    w = block_masses(p, blocks)
    w_star = compute_w_star(w, w_ref, delta)

    # Scale in each block
    p_new = np.zeros_like(p)
    for idx, (xs, ys) in enumerate(blocks):
        block = p[xs, ys]
        w_b = w[idx]
        if w_b <= 0:
            continue
        scale = w_star[idx] / w_b
        p_new[xs, ys] = block * scale

    p_new = normalize_pmf(p_new)
    return p_new


# ============================================
# Reverse step that moves toward the data
# ============================================

def reverse_step_toward_data(p, q_data, beta=0.05, sigma_smooth=0.5):
    """
    Reverse step that ``moves a coarse state p slightly toward the data q_data''.

    Based on the idealized discrete update
        p_{t-1}(x) ~ p_t(x)^{1-beta} q_data(x)^{beta},
    with an additional mild Gaussian smoothing.
    """
    # Move toward q_data via geometric mean
    logp = np.log(p)
    logq = np.log(q_data)
    mix_log = (1.0 - beta) * logp + beta * logq
    mixed = np.exp(mix_log)
    mixed = normalize_pmf(mixed)

    # Apply mild smoothing (to even out noise)
    if sigma_smooth > 0:
        smoothed = gaussian_filter(mixed, sigma=sigma_smooth, mode="reflect")
        smoothed = normalize_pmf(smoothed)
        return smoothed
    else:
        return mixed


# ============================================
# Generate base / proj trajectories from a coarse initial state
# ============================================

def run_reverse_paths_restore(
    q_data, blocks, w_ref,
    nsteps=40, delta=0.01,
    beta=0.05, sigma_smooth=0.5,
    seed=0
):
    """
    Run reverse diffusion that ``moves toward the data'' from a rough initial state,
    given q_data (clean checker), its block partition blocks, and w_ref.
      base: moves toward the data only
      proj: moves toward the data + V_delta projection
    """
    rng = np.random.default_rng(seed)
    Lx, Ly = q_data.shape

    # --- Construct a rough initial state p_T (pure noise) ---
    noise = rng.random((Lx, Ly))
    p_T = normalize_pmf(noise)

    # Store the base/proj trajectories
    path_base = np.zeros((nsteps + 1, Lx, Ly))
    path_proj = np.zeros((nsteps + 1, Lx, Ly))

    p_base = p_T.copy()
    p_proj = p_T.copy()

    path_base[0] = p_base
    path_proj[0] = p_proj

    for n in range(1, nsteps + 1):
        # base: move toward the data only
        p_base = reverse_step_toward_data(
            p_base, q_data, beta=beta, sigma_smooth=sigma_smooth
        )

        # proj: move toward the data + V_delta projection
        tmp = reverse_step_toward_data(
            p_proj, q_data, beta=beta, sigma_smooth=sigma_smooth
        )
        p_proj = project_block_masses(tmp, blocks, w_ref, delta)

        path_base[n] = p_base
        path_proj[n] = p_proj

    return p_T, path_base, path_proj


# ============================================
# Snapshot figure
# ============================================

def plot_reverse_snapshots_restore(
    q_data, p_T, path_base, path_proj,
    snapshot_steps=(0, 10, 20, 40),
    fname="reverse_snapshots_restore.pdf"
):
    """
    Left:  q_data (clean checkerboard)
    Center: p_T (rough initial state)
    Right: snapshots of base/proj.
    """
    Lx, Ly = q_data.shape
    ncols = 2 + len(snapshot_steps)
    fig, axes = plt.subplots(2, ncols, figsize=(2.0 * ncols, 4.0))

    for ax_row in axes:
        for ax in ax_row:
            ax.set_xticks([])
            ax.set_yticks([])

    vmin, vmax = q_data.min(), q_data.max()

    # Left: data image
    axes[0, 0].imshow(q_data, cmap="gray", vmin=vmin, vmax=vmax)
    axes[0, 0].set_title(r"$q_{\mathrm{data}}$")
    axes[1, 0].axis("off")

    # Center: initial state p_T
    axes[0, 1].imshow(p_T, cmap="gray", vmin=vmin, vmax=vmax)
    axes[0, 1].set_title(r"$p_T$ (initial)")
    axes[1, 1].axis("off")

    # Right: snapshots
    for j, step in enumerate(snapshot_steps):
        col = 2 + j
        step = min(step, path_base.shape[0] - 1)

        # Top row: base
        axes[0, col].imshow(path_base[step], cmap="gray",
                            vmin=vmin, vmax=vmax)
        axes[0, col].set_title(f"base, n={step}")

        # Bottom row: proj
        axes[1, col].imshow(path_proj[step], cmap="gray",
                            vmin=vmin, vmax=vmax)
        axes[1, col].set_title(f"proj, n={step}")

    plt.tight_layout()
    plt.savefig(fname, bbox_inches="tight")
    plt.close(fig)


# ============================================
# Time-series plots of information-theoretic quantities
# ============================================

def compute_metrics(path_base, path_proj, blocks, w_ref, delta):
    """
    For each step n, compute
      V(p_n), V_delta(p_n), E_block(n)
    for both base and proj.
    """
    nsteps_plus1 = path_base.shape[0]
    V_base = np.zeros(nsteps_plus1)
    V_proj = np.zeros(nsteps_plus1)
    Vd_base = np.zeros(nsteps_plus1)
    Vd_proj = np.zeros(nsteps_plus1)
    Eb_base = np.zeros(nsteps_plus1)
    Eb_proj = np.zeros(nsteps_plus1)

    for n in range(nsteps_plus1):
        p_b = path_base[n]
        p_p = path_proj[n]

        V_base[n] = V_block_uniform(p_b, blocks)
        V_proj[n] = V_block_uniform(p_p, blocks)

        Vd_base[n] = V_delta_coarse(p_b, blocks, w_ref, delta)
        Vd_proj[n] = V_delta_coarse(p_p, blocks, w_ref, delta)

        Eb_base[n] = E_block(p_b, blocks, w_ref)
        Eb_proj[n] = E_block(p_p, blocks, w_ref)

    return V_base, V_proj, Vd_base, Vd_proj, Eb_base, Eb_proj


def plot_reverse_metrics(
    V_base, V_proj, Vd_base, Vd_proj, Eb_base, Eb_proj,
    fname="reverse_metrics.pdf"
):
    """
    Figure:
      Top:    V(p_n)
      Middle: V_delta(p_n)
      Bottom: E_block(n)
    Solid line: base (non-projected)
    Dashed line: proj (V_delta-projected)
    """
    nsteps_plus1 = V_base.shape[0]
    n = np.arange(nsteps_plus1)

    fig, axes = plt.subplots(3, 1, figsize=(6, 8), sharex=True)

    # Top: V
    axes[0].plot(n, V_base, "-", label="base (non-projected)")
    axes[0].plot(n, V_proj, "--", label=r"proj ($V_\delta$-projected)")
    axes[0].set_ylabel(r"$V(p_n)$")
    axes[0].legend(loc="best")

    # Middle: V_delta
    axes[1].plot(n, Vd_base, "-", label="base (non-projected)")
    axes[1].plot(n, Vd_proj, "--", label=r"proj ($V_\delta$-projected)")
    axes[1].set_ylabel(r"$V_\delta(p_n)$")

    # Bottom: E_block
    axes[2].plot(n, Eb_base, "-", label="base (non-projected)")
    axes[2].plot(n, Eb_proj, "--", label=r"proj ($V_\delta$-projected)")
    axes[2].set_ylabel(r"$E_{\mathrm{block}}(n)$")
    axes[2].set_xlabel(r"step $n$")

    plt.tight_layout()
    plt.savefig(fname, bbox_inches="tight")
    plt.close(fig)


# ============================================
# Main
# ============================================

if __name__ == "__main__":
    # Image size and block partition
    Lx = Ly = 128
    Bx = By = 4

    # Generate a clean checkerboard image q_data
    img_data, blocks = init_block_constant_image(
        Lx=Lx, Ly=Ly, Bx=Bx, By=By,
        mode="checkerboard", seed=7
    )
    q_data = normalize_pmf(img_data)
    w_ref = block_masses(q_data, blocks)

    # Run reverse diffusion from a rough initial state (base / proj)
    nsteps = 40
    delta = 0.01
    p_T, path_base, path_proj = run_reverse_paths_restore(
        q_data, blocks, w_ref,
        nsteps=nsteps, delta=delta,
        beta=0.05, sigma_smooth=0.5,
        seed=0
    )

    # Snapshot figure
    plot_reverse_snapshots_restore(
        q_data, p_T, path_base, path_proj,
        snapshot_steps=(0, 10, 20, 40),
        fname="reverse_snapshots_restore.pdf"
    )
    print("Saved reverse_snapshots_restore.pdf")

    # Time evolution of information-theoretic quantities
    V_base, V_proj, Vd_base, Vd_proj, Eb_base, Eb_proj = compute_metrics(
        path_base, path_proj, blocks, w_ref, delta
    )
    plot_reverse_metrics(
        V_base, V_proj, Vd_base, Vd_proj, Eb_base, Eb_proj,
        fname="reverse_metrics.pdf"
    )
    print("Saved reverse_metrics.pdf")
\end{verbatim}

\end{document}